\documentclass{article} 
\usepackage{iclr2027_conference}
\usepackage{times}

\usepackage{amsmath,amsfonts,bm}

\def\eqref#1{equation~\ref{#1}}

\def\1{\bm{1}}

\DeclareMathAlphabet{\mathsfit}{\encodingdefault}{\sfdefault}{m}{sl}
\SetMathAlphabet{\mathsfit}{bold}{\encodingdefault}{\sfdefault}{bx}{n}

\usepackage{graphicx}
\usepackage{hyperref}
\usepackage{url}
\usepackage{subcaption}

\usepackage{booktabs}
\usepackage{multirow}
\usepackage[table]{xcolor}
\usepackage{graphicx}

\usepackage{algorithm}
\usepackage{algpseudocode}

\usepackage{xcolor}
\usepackage{pifont}
\usepackage{wrapfig}

\definecolor{gain}{RGB}{0,150,70}
\definecolor{loss}{RGB}{210,45,45}
\definecolor{checkgreen}{RGB}{46,125,50}
\definecolor{crossred}{RGB}{198,40,40}

\newcommand{\gain}[1]{\,{\tiny\textcolor{gain}{$\uparrow$#1}}}
\newcommand{\loss}[1]{\,{\tiny\textcolor{loss}{$\downarrow$#1}}}
\newcommand{\cmark}{\textcolor{checkgreen}{\ding{51}}}
\newcommand{\xmark}{\textcolor{crossred}{\ding{55}}}

\iclrfinalcopy

\title{Reason Through the Latent!\\Making Latent Visual Reasoning Necessary}

\author{Suhyeong Park$^{1,2}$, Junha Jung$^{1,2}$, Jaewoo Kang$^{1,2}$\thanks{Corresponding author.} \\
Korea University$^{1}$ \quad AIGEN Sciences$^{2}$ \\
\texttt{\{pshpulip22, goodjungjun, kangj\}@korea.ac.kr}
}

\begin{document}

\maketitle

\begin{abstract}
Latent visual reasoning aims to perform multimodal reasoning through hidden-state computation rather than explicit textual chains of thought. However, visual information being present in a latent state does not imply that the model actually relies on that state when producing its answer, especially when alternative image-conditioned paths remain available. We introduce \textbf{C}ausal \textbf{V}isual \textbf{R}ecurrent \textbf{R}easoning (CVRR), which preserves pretrained visual competence while making recurrent computation the required image-conditioned path to prediction. CVRR initializes recurrence from the question hidden state after the pretrained vision-language model has incorporated the image, then repeatedly updates this state while re-reading the same fixed visual evidence. Before decoding, visual states and the original multimodal KV cache are removed so that only the final recurrent state carries image-conditioned information to the answer. Across the $V^*$, MMVP, BLINK, and MME-RealWorld-Lite benchmarks, CVRR retains strong performance under this strict interface, while compatible latent reasoners fail to recover comparable visual competence even when retrained under the same constraint. Causal interventions further show that predictions remain sensitive to recurrent content when the question is held fixed, and that persistent visual evidence causally revises the recurrent trajectory. These results distinguish latent informativeness from latent computation that is actually used for prediction.\footnote{Our code is publicly available at \href{https://github.com/dmis-lab/CVRR}{https://github.com/dmis-lab/CVRR}.}
\end{abstract}

\section{Introduction}

Latent visual reasoning promises to let multimodal models reason through hidden states without verbalizing every intermediate step. Continuous latent reasoning has shown that intermediate computation need not be expressed as text~\citep{i2,i3}, and recent multimodal methods extend this idea to visual inputs through latent visual tokens or internal hidden states~\citep{i5,i6}. These methods differ in how latent states are constructed, updated, and coupled with visual evidence~\citep{i7}, with recent work exploring recurrent updates, hybrid text--latent trajectories, and repeated interaction with the image. Such approaches are particularly appealing for visual reasoning, where relevant evidence may be spatially distributed and difficult to express faithfully through intermediate text. Latent visual reasoning therefore offers a natural substrate for maintaining and refining visual information across multi-step inference.

However, a latent state containing task-relevant visual information is not necessarily one that the model actually uses to determine its answer. To test whether existing latent visual reasoners depend on their proposed latent computation, we intervene on their latent states while leaving the original multimodal answer context intact. For several methods, predictions remain largely unchanged, showing that the proposed latent content is not behaviorally necessary when alternative image-conditioned routes remain available. This exposes a central ambiguity in latent visual reasoning: an internal state can be informative without being causally required for prediction. We do not argue that every useful latent reasoner must eliminate all redundant visual paths. Rather, when the scientific claim is that prediction depends on a proposed latent computation, latent informativeness alone is insufficient evidence that the model actually uses that computation. Our goal is therefore not simply to make latent states informative, but to make prediction depend on computation through them without sacrificing the visual competence formed by pretraining.

This perspective suggests two design requirements. The latent computation should begin from a representation that preserves pretrained visual competence, and it should become the required route through which that competence reaches the answer. We introduce \textbf{C}ausal \textbf{V}isual \textbf{R}ecurrent \textbf{R}easoning (CVRR) to satisfy both. CVRR initializes its latent state from the question-token hidden state after the pretrained vision-language model (VLM) has incorporated the image. It then reuses a single decoder layer as a shared recurrent transition, repeatedly updating the question state while re-reading fixed visual evidence. After the final recurrent step, visual rows and the original multimodal KV cache are removed, leaving the recurrent question state as the sole image-conditioned path to prediction. Path necessity is deliberately enforced by construction rather than discovered post hoc. The nontrivial question is whether this constraint can be imposed without destroying pretrained visual competence and whether computation along the required path performs prediction-relevant visual refinement. CVRR is designed to address both.

Experiments across $V^*$~\citep{e2}, MMVP~\citep{e3}, BLINK~\citep{e4}, and MME-RealWorld-Lite~\citep{e5} show that this path constraint can be imposed while retaining strong visual reasoning performance. CVRR reaches $81.2\%$ on $V^*$, $52.7\%$ MMVP pair accuracy, and $55.2\%$ on BLINK overall under the strict no-bypass interface, while compatible latent visual reasoners fail to recover comparable competence even when retrained under the same constraint. Our analyses show that this performance does not arise from simply routing a pretrained representation through a required bottleneck. Retraining without recurrent visual re-reading reduces $V^*$ accuracy by $12.0$ points and MMVP pair accuracy by $34.7$ points despite retaining native $h_1$ and the learned transition, while removing the learned transition also degrades performance. Causal interventions show that persistent visual evidence reshapes the recurrent trajectory and that predictions remain tied to image-conditioned recurrent content when the question is controlled. Together, these results show that CVRR makes latent computation necessary for prediction while performing learned, prediction-relevant visual refinement along that required path. Our contributions are:
\begin{enumerate}
    \item We recast latent-state reliance in visual reasoning as a causal path-design problem, distinguishing latent informativeness from path necessity.

    \item We introduce CVRR, which preserves pretrained competence through image-conditioned initialization while making recurrent latent computation necessary for prediction.

    \item We show through ablations and causal interventions that CVRR performs prediction-relevant visual refinement through learned updates and persistent visual re-reading.
\end{enumerate}
\section{Preliminary Experiments}

\begin{figure}[hbt!]
    \centering

    \begin{subfigure}[t]{0.51\linewidth}
        \centering
        \includegraphics[width=\linewidth]{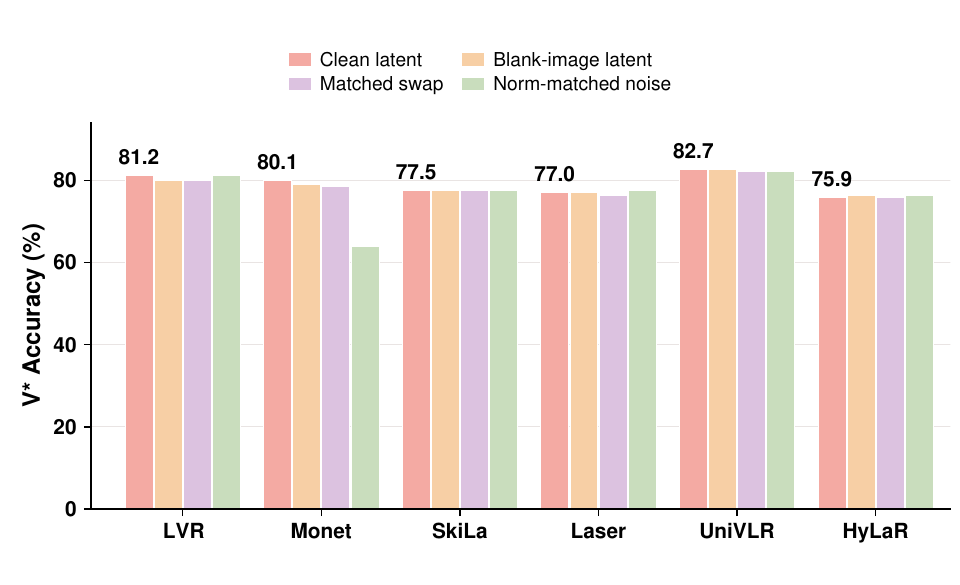}
        \caption{\textit{Latent content is often nonessential}}
        \label{fig:prelim_latent_replacement}
    \end{subfigure}
    \hfill
    \begin{subfigure}[t]{0.48\linewidth}
        \centering
        \includegraphics[width=\linewidth]{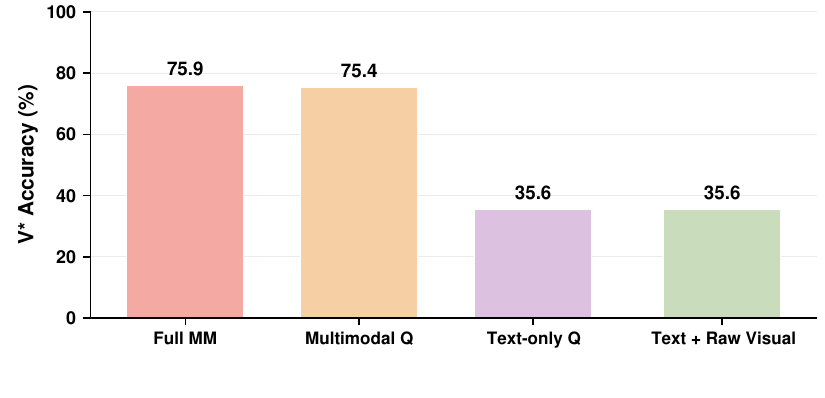}
        \caption{\textit{Multimodal question states preserve competence}}
        \label{fig:prelim_image_kv}
    \end{subfigure}

    \caption{
    Preliminary diagnostics motivating CVRR.
    (a) Accuracy under blank-image, matched-state, and norm-matched-noise replacements relative to clean latent states.
    (b) Accuracy from the full multimodal state (\textit{Full MM}), its question rows (\textit{Multimodal Q}), text-only question rows (\textit{Text-only Q}), and text-only rows with raw visual features (\textit{Text+Raw Visual}).
    Panel (b) uses the frozen pretrained Qwen2.5-VL-7B-Instruct before CVRR adaptation.
    }
    \label{fig:prelim}
\end{figure}

We first establish two observations that motivate CVRR: existing latent visual reasoners need not depend strongly on their proposed latent content, and pretrained visual competence is already integrated into the native multimodal question state.

\subsection{Weak Behavioral Reliance on Latent Content}
\label{sec:prelim_1}

We first ask whether existing latent visual reasoners actually depend on their proposed latent states. For each method introduced in Appendix~\ref{app:related_work_LVR}, we replace the latent content with a blank-image latent, a matched latent from an example with a different answer, or row-norm-matched noise, while leaving the original multimodal answer context intact. We compare each intervention with the paired clean prediction from the unaltered latent state. Small changes in accuracy therefore indicate weak behavioral dependence on the proposed latent content.

As shown in Figure~\ref{fig:prelim_latent_replacement}, several methods remain close to their clean accuracy even after substantial changes to the latent state. For example, UniVLR changes by at most $0.5$ percentage point across all three replacements. These interventions are not strict no-bypass evaluations because the original multimodal answer context remains available. Instead, they ask whether the standard inference path actually requires the proposed latent content. Stable predictions under replacement show that the answer can be sustained without strong dependence on that content. Thus, strong task performance does not by itself establish that prediction depends on the proposed latent computation.

\subsection{Multimodal Question States Preserve Visual Competence}
\label{sec:prelim_2}

Removing alternative answer paths raises a complementary question: which representation should carry pretrained visual competence once those paths are unavailable? We compare four states formed at an intermediate layer of the frozen pretrained VLM: the full multimodal hidden state containing visual and question tokens (\textit{Full MM}), the question rows from the same multimodal forward pass (\textit{Multimodal Q}), the question rows from a text-only pass (\textit{Text-only Q}), and the text-only question rows concatenated with raw visual features (\textit{Text+Raw Visual}). Each state is passed alone to the same frozen upper decoder. As shown in Figure~\ref{fig:prelim_image_kv}, \textit{Full MM} reaches $75.9\%$ and \textit{Multimodal Q} reaches $75.4\%$, whereas both text-only conditions achieve only $35.6\%$.

The $0.5$ percentage point gap between \textit{Full MM} and \textit{Multimodal Q} indicates that nearly all of the visual competence available from the full multimodal state is already carried by the question representation at this layer. In contrast, attaching raw visual features to a text-only state does not recover this competence. Together with the weak latent reliance above, these findings motivate two requirements for latent visual reasoning: the latent path should begin from the native multimodal question representation, and alternative image-conditioned answer paths should be removed so that prediction actually depends on computation through that state.
\begin{figure*}[t!]
    \centering
    \includegraphics[width=\textwidth]{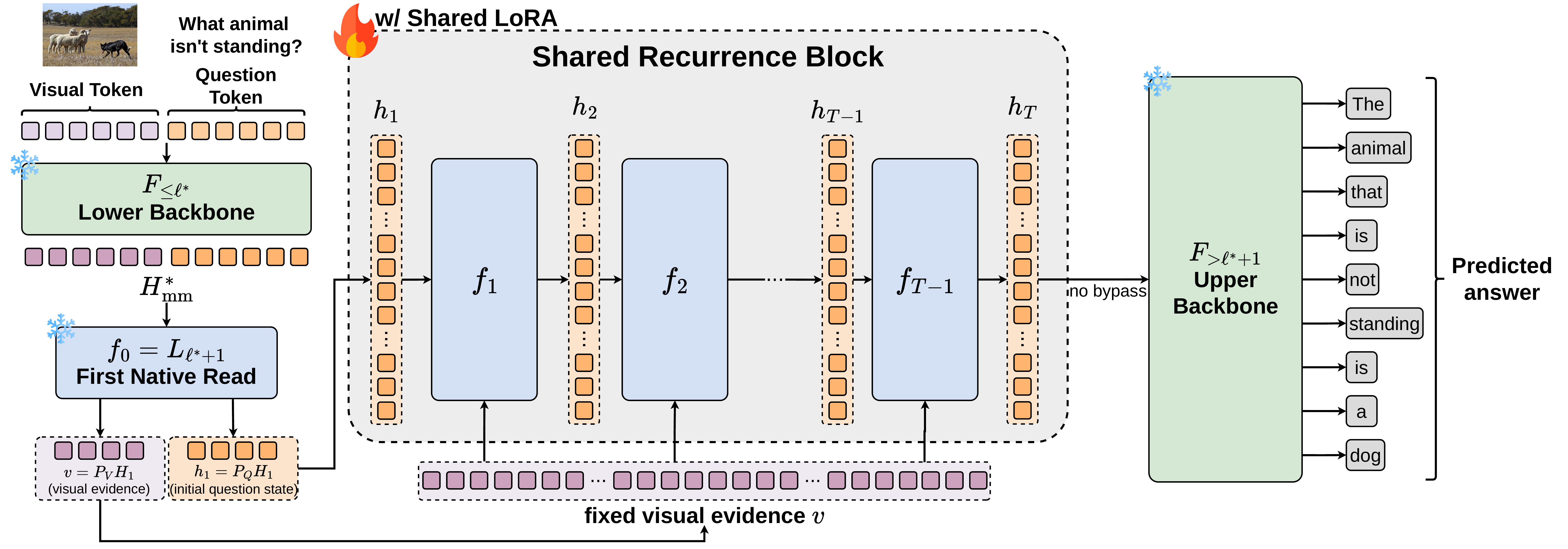}
    \caption{
    Overview of CVRR. The native multimodal forward forms the image-conditioned question state $h_1$ and persistent visual evidence $v$ at the causal visual-read boundary. A shared adapted layer repeatedly updates $h_t$ while re-reading the same fixed $v$. After $T$ steps, visual states and multimodal caches are discarded, and only $h_T$ enters the frozen upper backbone for answer prediction.
    }
    \label{fig:CVRR}
\end{figure*}

\section{Method}
\label{sec:method}

The preliminary results suggest that latent visual reasoning requires two properties simultaneously. The latent path should preserve the visual competence already formed by the pretrained VLM, and prediction should actually depend on computation through that path. CVRR is designed around these requirements. It begins recurrence from a native image-conditioned question representation, repeatedly refines this state while re-reading persistent visual evidence, and removes all alternative image-conditioned routes before answer decoding. The resulting recurrent state therefore carries pretrained visual competence while becoming the required visual path to prediction. Figure~\ref{fig:CVRR} summarizes the architecture, and Appendix~\ref{app:algorithm} provides the complete forward procedure.

\subsection{Causal Visual-Read Boundary}

The recurrent path should not begin from an arbitrary hidden state. It should start after substantial visual information has entered the question representation, while visual states still retain downstream influence that recurrence can re-read. We locate this transition using layer-wise activation patching on the frozen base VLM~\citep{m1,m2}. We construct contrastive pairs $x=(I,q,y)$ and $x'=(I',q,y')$ that share the same question but contain different images and yield different answers. At each candidate layer, we replace the visual-token activations of one example with those from its paired example and measure how much this reduces the preference for its original answer. Let $M_x$ denote the log-probability margin favoring the answer of $x$ over that of $x'$, with $M_{x'}$ defined in the opposite direction, and let $P_{\ell}(x\leftarrow x')$ denote patching the visual activations of $x$ with those of $x'$ at layer $\ell$. We measure the bidirectional effect as
\begin{equation}
\begin{aligned}
m_i(\ell)
=
\frac{1}{2}\Big[
&
M_x(x)-M_x\!\left(P_{\ell}(x\leftarrow x')\right)
\\
+{}&
M_{x'}(x')-M_{x'}\!\left(P_{\ell}(x'\leftarrow x)\right)
\Big],
\qquad
\widehat m_i(\ell)=\frac{m_i(\ell)}{m_i(0)},
\end{aligned}
\label{eq:causal_boundary}
\end{equation}
where larger normalized effects indicate that visual states at that layer still exert substantial downstream influence. We aggregate these profiles across contrastive pairs and use the layer preceding the first sustained decline as the candidate recurrent boundary.

For Qwen2.5-VL-7B~\citep{e7}, this procedure yields $\ell^*=20$, and the subsequent layer $f_0=L_{21}$ initializes recurrence. At this boundary, the frozen lower backbone produces
\begin{equation}
H_{\mathrm{mm}}^*
=
F_{\leq\ell^*}^{\mathrm{mm}}(I,q),
\qquad
Q_{\mathrm{txt}}^*
=
F_{\leq\ell^*}^{\mathrm{txt}}(q).
\label{eq:boundary_states}
\end{equation}
The multimodal state supplies the native image-conditioned representation from which recurrence begins, whereas the text-only state is reserved for image-free decoder context. Their question rows preserve the same token identity, ordering, and masking structure. The multimodal forward is terminated after $f_0$, and no later multimodal hidden state or prefix KV cache is retained for answer decoding. Appendix~\ref{app:boundary_localization} further shows that this region jointly provides strong question-state sufficiency and residual visual influence and independently validates nearby boundaries.

\subsection{Persistent Visual Recurrence}

A competent image-conditioned initialization is necessary for preserving pretrained visual ability, but initialization alone does not provide further visual computation. CVRR therefore keeps visual evidence available throughout recurrence while allowing the question state to evolve. Reusing shared Transformer parameters across depth follows prior work on universal, looped, and recurrent-depth Transformers~\citep{loop1,loop2,loop3}. Unlike these approaches, CVRR uses recurrence to repeatedly re-read persistent visual evidence from a native multimodal state while making the resulting recurrent state the required image-conditioned path to prediction. Given the multimodal boundary state, we first apply the subsequent decoder layer $f_0=L_{\ell^*+1}$ once in its pretrained form, without the low-rank adapter used in later recurrent steps:
\begin{equation}
H_1=f_0(H_{\mathrm{mm}}^*),
\qquad
h_1=P_QH_1,
\qquad
v=P_VH_1.
\label{eq:initial_states}
\end{equation}
Here, $P_Q$ and $P_V$ select the question-token and visual-token rows. The question rows $h_1$ form the initial recurrent state, while the visual rows $v$ provide persistent evidence throughout recurrence. Because both originate from the same native multimodal forward pass, recurrence begins from the visual computation already established by the pretrained model rather than attempting to reconstruct it from raw visual features.

At each subsequent step, we reconstruct the native multimodal scaffold by replacing only the question rows with the previous recurrent state while keeping the visual rows fixed. Each transition therefore receives an evolving question state together with the same visual evidence:
\begin{equation}
\begin{aligned}
H_t
&=
\operatorname{Replace}_{Q}(H_1,h_{t-1}),
\qquad
P_VH_t=v,
\\[0.35em]
\widetilde h_t
&=
P_Qf_{\phi}(H_t),
\\[0.35em]
h_t
&=
(1-\beta)h_{t-1}
+
\beta\widetilde h_t,
\qquad
t=2,\ldots,T.
\end{aligned}
\label{eq:recurrence}
\end{equation}
Visual re-reading occurs through the native causal self-attention of the shared decoder layer without introducing a separate cross-attention module. The original token ordering, causal mask, and positional structure are preserved, and all recurrent steps share the same layer and low-rank parameters. Thus, $h_t$ is repeatedly revised against persistent visual evidence rather than transformed in isolation.

\subsection{Strict Causal Decoder Interface and Training Objective}

Even a visually grounded recurrent state need not be behaviorally necessary if the decoder can still access the original multimodal context. CVRR therefore removes every alternative image-conditioned route before answer decoding. Below the recurrent boundary, autoregressive answer tokens use only image-free prefix caches $\mathcal K_{\mathrm{txt}}$ constructed from the text-only question path. Above the boundary, the final recurrent state $h_T$ serves as the question-prefix representation for the frozen upper decoder. All visual rows and original multimodal KV caches are discarded before decoding. The text-only caches provide the lower-layer linguistic context required for autoregressive processing but contain no image-conditioned information. Consequently, $h_T$ is the sole source of image-conditioned information available to the answer decoder.

The backbone and upper decoder remain frozen, and only the shared low-rank parameters $\phi$ are optimized. We use no latent targets, teacher trajectories, or reconstruction objectives to encourage reliance on the recurrent state. Instead, path reliance follows from the decoder interface itself. For a minibatch $\mathcal B$ and valid answer-token set $\mathcal A_i$, we minimize answer-token cross-entropy,
\begin{equation}
\mathcal{L}_{\mathrm{ans}}
=
-\frac{1}{|\mathcal B|}
\sum_{i\in\mathcal B}
\frac{1}{|\mathcal A_i|}
\sum_{j\in\mathcal A_i}
\log
p_{\phi}
\left(
y_{i,j}
\mid
y_{i,<j},
h_{T,i},
\mathcal K_{\mathrm{txt},i}
\right).
\label{eq:loss}
\end{equation}
Prompt and padding positions are excluded from the loss, and token losses are averaged within each example and then across the minibatch. The same recurrent trajectory and strict decoder interface are used at inference time.
\section{Experiments}

\subsection{Datasets \& Metrics}

We train on Visual CoT~\citep{e1}, containing 438K QA pairs with answer-relevant bounding-boxes. We evaluate on four VQA benchmarks: $V^*$~\citep{e2}, MMVP~\citep{e3}, BLINK~\citep{e4}, and MME-RealWorld-Lite\footnote{\url{https://huggingface.co/datasets/yifanzhang114/MME-RealWorld-Lite}}~\citep{e5}. $V^*$ contains 191 high-resolution examples, while MMVP contains 300 questions over 150 visually similar image pairs. From BLINK, we report the Counting, IQ-Test, Jigsaw, Relative Reflectance, and Spatial Relation subsets following LVR~\citep{i6}, together with overall accuracy across the benchmark. We report accuracy for all benchmarks: overall accuracy and the Direct Attributes (D.A.) and Relative Position (R.P.) splits for $V^*$; item and pair accuracy for MMVP, where both questions in a pair must be correct; per-subset and overall accuracy for BLINK; and overall accuracy for MME-RealWorld-Lite.

\subsection{Baselines}

Our comparisons are designed to separate visual competence from causal reliance on the latent path. As general-purpose reference models, we evaluate GPT-5~\citep{e6}, Qwen2.5-VL-7B-Instruct, and Qwen3-VL-8B-Thinking~\citep{e8} with chain-of-thought (CoT) reasoning. Visual reasoning models include VL-Rethinker~\citep{r1}, DeepEyes~\citep{r2}, and PixelReasoner~\citep{r3}, while latent visual reasoning methods include LVR, Monet~\citep{r4}, SkiLa~\citep{r5}, Laser~\citep{r6}, UniVLR~\citep{r7}, HyLaR~\citep{r8}, and PEARL~\citep{r9}. Qwen2.5-VL-7B-Instruct serves as the pretrained backbone reference for CVRR. To distinguish recurrent computation from ordinary supervised adaptation, we additionally train a full-multimodal SFT control on the same Visual CoT data while retaining the standard multimodal answer path.

The most direct comparison tests whether removing bypasses alone is sufficient. We retrain compatible latent reasoning baselines under the same strict decoder interface while preserving their original latent mechanisms and training objectives. These no-bypass variants use the same Visual CoT data and common SFT settings as CVRR, but visual states and multimodal prefix caches are removed before answer decoding. This comparison isolates the central design question: whether a method can make its latent path necessary without losing the visual competence of the pretrained model. Implementation and evaluation details are provided in Appendix~\ref{app:implementation}.

\begin{table*}[t]
\centering
\caption{
Results across $V^*$, MMVP, BLINK, and MME-RealWorld-Lite.
\textbf{Bold} and \underline{underline} indicate the best and second-best within each model group.
Colored arrows show point changes from the corresponding pretrained backbone.
Indented \textit{w/ SFT} rows denote full-multimodal controls trained on the same data without recurrent reasoning.
The shaded row denotes CVRR.
}
\label{tab:main_results}

\setlength{\tabcolsep}{2.0pt}
\renewcommand{\arraystretch}{1.08}
\small
\resizebox{\textwidth}{!}{
\begin{tabular}{@{}lcccccccccccc@{}}
\toprule
\multirow{2}{*}{Method}
& \multicolumn{3}{c}{$V^*$}
& \multicolumn{2}{c}{MMVP}
& \multicolumn{6}{c}{BLINK}
& \multicolumn{1}{c}{MME-RW-Lite} \\
\cmidrule(lr){2-4}
\cmidrule(lr){5-6}
\cmidrule(lr){7-12}
\cmidrule(l){13-13}
& Overall & D.A. & R.P.
& Item & Pair
& Overall & Counting & IQ-Test & Jigsaw & Rel. Reflect & Spatial Rel.
& Overall \\
\midrule

\multicolumn{13}{l}{\textit{General-Purpose VLMs w/ CoT}} \\

GPT-5
& 73.3
& 70.4
& 77.6
& \textbf{87.7}
& \textbf{77.3}
& \textbf{74.0}
& \textbf{78.3}
& \textbf{35.3}
& \textbf{78.0}
& \textbf{68.7}
& \underline{89.5}
& \textbf{57.8} \\

Qwen3-VL-8B-Thinking
& 78.5
& \underline{81.7}
& 73.7
& 52.3
& 20.7
& 52.8
& 65.8
& 18.7
& 44.7
& \underline{57.5}
& 55.9
& 46.7 \\

Qwen2.5-VL-7B-Instruct
& 79.6
& 79.1
& \textbf{80.3}
& 70.3
& 47.3
& 55.0
& \underline{74.2}
& \underline{30.0}
& \underline{67.3}
& 41.0
& 84.6
& \underline{50.8} \\

\quad \textit{w/ SFT}
& 80.6\gain{1.0}
& 80.9\gain{1.8}
& \textbf{80.3}
& 74.7\gain{4.4}
& 52.0\gain{4.7}
& 53.0\loss{2.0}
& 70.8\loss{3.4}
& 24.0\loss{6.0}
& 59.3\loss{8.0}
& 34.3\loss{6.7}
& \textbf{90.9}\gain{6.3}
& 44.7\loss{6.1} \\



\midrule
\multicolumn{13}{l}{\textit{Visual Reasoning}} \\

VL-Rethinker
& \underline{81.7}\gain{2.1}
& \underline{80.9}\gain{1.8}
& \underline{82.9}\gain{2.6}
& \textbf{74.3}\gain{4.0}
& \textbf{52.0}\gain{4.7}
& \textbf{54.5}\loss{0.5}
& \underline{68.3}\loss{5.9}
& \underline{20.7}\loss{9.3}
& \textbf{66.0}\loss{1.3}
& \underline{41.0}
& \textbf{90.2}\gain{5.6}
& \underline{50.9}\gain{0.1} \\

DeepEyes
& 81.2\gain{1.6}
& \textbf{81.7}\gain{2.6}
& 80.3
& 69.7\loss{0.6}
& 43.3\loss{4.0}
& \underline{52.7}\loss{2.3}
& \textbf{70.8}\loss{3.4}
& \textbf{26.7}\loss{3.3}
& 54.7\loss{12.6}
& 37.3\loss{3.7}
& 81.8\loss{2.8}
& \textbf{54.1}\gain{3.3} \\

PixelReasoner
& \textbf{82.7}\gain{3.1}
& \textbf{81.7}\gain{2.6}
& \textbf{84.2}\gain{3.9}
& \underline{73.3}\gain{3.0}
& \underline{48.0}\gain{0.7}
& 51.1\loss{3.9}
& 67.5\loss{6.7}
& \textbf{26.7}\loss{3.3}
& \underline{64.0}\loss{3.3}
& \textbf{44.8}\gain{3.8}
& \underline{83.2}\loss{1.4}
& 50.6\loss{0.2} \\

\midrule
\multicolumn{13}{l}{\textit{Latent Visual Reasoning}} \\

LVR-7B
& \underline{81.2}\gain{1.6}
& \underline{83.5}\gain{4.4}
& \textbf{77.6}\loss{2.7}
& 71.0\gain{0.7}
& 44.7\loss{2.6}
& 51.8\loss{3.2}
& \textbf{70.8}\loss{3.4}
& \textbf{28.0}\loss{2.0}
& 58.0\loss{9.3}
& 38.8\loss{2.2}
& \underline{89.5}\gain{4.9}
& \underline{51.3}\gain{0.5} \\

Monet-7B
& 79.1\loss{0.5}
& 80.9\gain{1.8}
& \underline{76.3}\loss{4.0}
& 70.7\gain{0.4}
& 46.0\loss{1.3}
& 46.3\loss{8.7}
& 60.8\loss{13.4}
& 17.3\loss{12.7}
& 50.7\loss{16.6}
& 41.8\gain{0.8}
& 79.0\loss{5.6}
& 50.1\loss{0.7} \\

SkiLa-7B
& 77.5\loss{2.1}
& 79.1
& 75.0\loss{5.3}
& 71.7\gain{1.4}
& 47.3
& 52.8\loss{2.2}
& 67.5\loss{6.7}
& 18.7\loss{11.3}
& 64.0\loss{3.3}
& 38.8\loss{2.2}
& \underline{89.5}\gain{4.9}
& 44.7\loss{6.1} \\

Laser-7B
& 77.5\loss{2.1}
& 82.6\gain{3.5}
& 69.7\loss{10.6}
& 64.7\loss{5.6}
& 32.7\loss{14.6}
& \underline{53.4}\loss{1.6}
& 63.3\loss{10.9}
& \underline{24.7}\loss{5.3}
& 63.3\loss{4.0}
& \underline{44.8}\gain{3.8}
& 83.2\loss{1.4}
& 50.8 \\

UniVLR-7B
& \textbf{82.7}\gain{3.1}
& \textbf{87.8}\gain{8.7}
& 75.0\loss{5.3}
& 70.0\loss{0.3}
& 42.7\loss{4.6}
& 50.2\loss{4.8}
& 66.7\loss{7.5}
& 22.0\loss{8.0}
& 42.0\loss{25.3}
& 42.5\gain{1.5}
& 84.6
& \textbf{51.8}\gain{1.0} \\

HyLaR-7B
& 75.9\loss{3.7}
& 77.4\loss{1.7}
& 73.7\loss{6.6}
& \underline{74.3}\gain{4.0}
& \underline{51.3}\gain{4.0}
& \underline{53.4}\loss{1.6}
& \underline{68.3}\loss{5.9}
& 20.0\loss{10.0}
& \underline{64.7}\loss{2.6}
& 43.3\gain{2.3}
& \textbf{90.2}\gain{5.6}
& 44.9\loss{5.9} \\

PEARL
& 80.6\gain{1.0}
& 82.6\gain{3.5}
& \textbf{77.6}\loss{2.7}
& \textbf{75.7}\gain{5.4}
& \textbf{54.7}\gain{7.4}
& \textbf{56.7}\gain{1.7}
& \underline{68.3}\loss{5.9}
& \underline{24.7}\loss{5.3}
& \textbf{67.3}
& \textbf{50.0}\gain{9.0}
& 88.8\gain{4.2}
& 47.6\loss{3.2} \\

\midrule
\multicolumn{13}{l}{\textit{Latent Visual Reasoning w/o Bypass}} \\

LVR-7B
& 35.6\loss{44.0}
& 24.4\loss{54.7}
& 52.6\loss{27.7}
& 50.0\loss{20.3}
& 0.7\loss{46.6}
& 37.9\loss{17.1}
& 39.2\loss{35.0}
& 20.0\loss{10.0}
& 52.7\loss{14.6}
& 26.9\loss{14.1}
& 53.9\loss{30.7}
& 26.2\loss{24.6} \\

Monet-7B
& 36.1\loss{43.5}
& 25.2\loss{53.9}
& 52.6\loss{27.7}
& \underline{50.7}\loss{19.6}
& 2.0\loss{45.3}
& 38.4\loss{16.6}
& \underline{45.0}\loss{29.2}
& \underline{24.7}\loss{5.3}
& 52.7\loss{14.6}
& 26.9\loss{14.1}
& 55.9\loss{28.7}
& \underline{33.1}\loss{17.7} \\

SkiLa-7B
& 38.7\loss{40.9}
& \underline{28.7}\loss{50.4}
& 54.0\loss{26.3}
& 29.3\loss{41.0}
& 0.0\loss{47.3}
& \underline{38.8}\loss{16.2}
& 40.0\loss{34.2}
& 24.0\loss{6.0}
& 47.3\loss{20.0}
& \underline{40.3}\loss{0.7}
& 51.1\loss{33.5}
& 15.6\loss{35.2} \\

Laser-7B
& 37.2\loss{42.4}
& 27.0\loss{52.1}
& 52.6\loss{27.7}
& 50.0\loss{20.3}
& 0.7\loss{46.6}
& 37.8\loss{17.2}
& 40.8\loss{33.4}
& 19.3\loss{10.7}
& 47.3\loss{20.0}
& 26.9\loss{14.1}
& 57.3\loss{27.3}
& 31.6\loss{19.2} \\

UniVLR-7B
& 37.7\loss{41.9}
& 27.8\loss{51.3}
& 52.6\loss{27.7}
& 49.3\loss{21.0}
& 0.0\loss{47.3}
& 38.4\loss{16.6}
& 40.0\loss{34.2}
& 20.0\loss{10.0}
& 52.7\loss{14.6}
& 26.9\loss{14.1}
& \underline{58.7}\loss{25.9}
& 27.3\loss{23.5} \\

HyLaR-7B
& \underline{39.8}\loss{39.8}
& 26.1\loss{53.0}
& \underline{60.5}\loss{19.8}
& 42.3\loss{28.0}
& \underline{2.7}\loss{44.6}
& 33.7\loss{21.3}
& 36.7\loss{37.5}
& 6.7\loss{23.3}
& \underline{53.3}\loss{14.0}
& 32.8\loss{8.2}
& 37.8\loss{46.8}
& 32.6\loss{18.2} \\

\rowcolor{gray!20}
CVRR
& \textbf{81.2}\gain{1.6}
& \textbf{82.6}\gain{3.5}
& \textbf{79.0}\loss{1.3}
& \textbf{74.7}\gain{4.4}
& \textbf{52.7}\gain{5.4}
& \textbf{55.2}\gain{0.2}
& \textbf{70.0}\loss{4.2}
& \textbf{28.0}\loss{2.0}
& \textbf{62.7}\loss{4.6}
& \textbf{42.5}\gain{1.5}
& \textbf{90.9}\gain{6.3}
& \textbf{46.7}\loss{4.1} \\
\bottomrule
\end{tabular}
}
\end{table*}

\subsection{Main Results}

The central comparison in Table~\ref{tab:main_results} is not absolute benchmark rank, but whether visual competence survives once the answer is required to depend on the latent path. CVRR retains this competence under the strict no-bypass interface, reaching $81.2\%$ on $V^*$, $52.7\%$ MMVP pair accuracy, and $55.2\%$ on BLINK overall. In contrast, latent reasoning baselines retrained under the same interface reach at most $39.8\%$ on $V^*$, $2.7\%$ MMVP pair accuracy, $38.8\%$ on BLINK, and $33.1\%$ on MME-RealWorld-Lite. The contrast shows that simply forcing prediction through a latent state is not enough. The path must also begin from a representation that preserves the visual computation established by pretraining, which CVRR obtains through native multimodal initialization.

The full-multimodal SFT control provides a complementary check that these results are not merely an effect of training on Visual CoT. CVRR matches or exceeds this control on $V^*$, MMVP, and BLINK despite removing the standard multimodal answer path.
On MME-RealWorld-Lite, CVRR drops $4.1$ points from its pretrained backbone, compared with a $6.1$ point drop for the SFT control. Thus, the strict recurrent interface does not introduce a larger degradation than ordinary supervised adaptation on this benchmark. Taken together, the results support the intended claim: latent computation can be made behaviorally necessary while retaining much of the visual competence that makes it useful.
\section{Analysis}
\label{sec:analysis}

The results show that CVRR preserves visual competence while making the recurrent state the required image-conditioned path to prediction. Path necessity alone, however, does not reveal what computation occurs along that path. We therefore examine whether learned recurrence improves on native initialization, whether persistent visual evidence causally shapes the trajectory, whether sensitivity to $h_T$ reflects image-conditioned content, and how visual influence changes across steps.

\subsection{What Does Recurrence Add Beyond $h_1$?}
\label{sec:component_ablation}

\begin{table*}[hbt!]
\centering
\caption{
Component analysis on $V^*$ ($n=191$) and MMVP ($n=300$).
$\Delta$ denotes the percentage-point change from Full CVRR.
Variants without native initialization replace $h_1$ with the text-only anchor $B$.
Component ablations use $T=4$ unless otherwise indicated.
}
\setlength{\tabcolsep}{3.0pt}
\resizebox{\textwidth}{!}{%
\begin{tabular}{lccccccccccc}
\toprule
& \multicolumn{3}{c}{Components}
& \multicolumn{4}{c}{$V^*$}
& \multicolumn{4}{c}{MMVP} \\
\cmidrule(lr){2-4}
\cmidrule(lr){5-8}
\cmidrule(lr){9-12}

Variant
& Native $h_1$
& Visual Re-read $v$
& Transition $f_{\phi}$
& Overall
& D.A.
& R.P.
& $\Delta$
& Item
& Pair
& $\Delta$ Item
& $\Delta$ Pair \\
\midrule

\textbf{Full CVRR ($T=4$)}
& \cmark
& \cmark
& \cmark
& \textbf{81.2}
& \textbf{82.6}
& \textbf{79.0}
& --
& \textbf{74.7}
& \textbf{52.7}
& --
& -- \\

\midrule
\multicolumn{12}{l}{\textit{Recurrent depth controls}} \\

$T=2$
& \cmark
& \cmark
& \cmark
& 78.5
& 80.0
& 76.3
& -2.6
& 74.0
& 50.7
& -0.7
& -2.0 \\

$T=1$
& \cmark
& \xmark
& \xmark
& 75.4
& 79.1
& 69.7
& -5.8
& 60.7
& 22.7
& -14.0
& -30.0 \\

\midrule
\multicolumn{12}{l}{\textit{Component ablations}} \\

w/o Visual Re-read
& \cmark
& \xmark
& \cmark
& 69.1
& 71.3
& 65.8
& -12.0
& 56.7
& 18.0
& -18.0
& -34.7 \\

w/o Native MM Init ($h_1\!\rightarrow\!B$)
& \xmark
& \cmark
& \cmark
& 37.2
& 27.0
& 52.6
& -44.0
& 50.7
& 2.0
& -24.0
& -50.7 \\

w/o Learned Transition ($f_{\phi}$)
& \cmark
& \cmark
& \xmark
& 76.4
& 80.0
& 71.1
& -4.7
& 65.7
& 32.7
& -9.0
& -20.0 \\

w/o Entire Visual Path
& \xmark
& \xmark
& \cmark
& 37.2
& 27.0
& 52.6
& -44.0
& 50.0
& 0.0
& -24.7
& -52.7 \\

\bottomrule
\end{tabular}%
}
\label{tab:component_ablation}
\end{table*}

Table~\ref{tab:component_ablation} separates the roles of native multimodal initialization and recurrent refinement. Here, $B$ denotes the text-only question state obtained by passing the same question without the image through the frozen lower backbone and native layer $f_0$, aligning it with $h_1$ in depth and token structure while removing image-conditioned content. Replacing $h_1$ with $B$ reduces $V^*$ accuracy from $81.2\%$ to $37.2\%$ and MMVP pair accuracy from $52.7\%$ to $2.0\%$ despite retaining visual evidence $v$ and a retrained $f_{\phi}$. On $V^*$, this matches removing the entire visual path and falls within the $35$--$40\%$ range of the no-bypass baselines in Table~\ref{tab:main_results}. Thus, persistent visual access does not reconstruct the competence already integrated into the native multimodal question state. The role of $h_1$ is therefore to carry the pretrained model's native image-conditioned representation into the required path.

Given this starting state, we isolate what recurrent computation contributes. All trainable component ablations are independently retrained under the same recipe as Full CVRR. Decoding $h_1$ at $T=1$ reaches $75.4\%$ on $V^*$ but only $22.7\%$ MMVP pair accuracy, showing that native initialization alone is insufficient. Removing visual re-reading while retaining native $h_1$, the learned transition, and $T=4$ reduces $V^*$ accuracy by $12.0$ points and MMVP pair accuracy by $34.7$ points. Thus, CVRR does not merely pass a competent $h_1$ through trainable latent transformations; persistent visual evidence provides prediction-relevant information during recurrence. Conversely, removing the learned transition reduces performance by $4.7$ and $20.0$ points, showing that visual access alone is also insufficient. The trained $T=2$ control reaches $78.5\%$ on $V^*$ and $50.7\%$ MMVP pair accuracy, while full CVRR reaches $81.2\%$ and $52.7\%$. Together, these results support CVRR's central design: pretrained visual competence enters through $h_1$ and is actively refined by learned computation that repeatedly re-reads visual evidence.

\subsection{Causal Visual Re-reading}
\label{sec:causal_reliance}

\begin{figure}[hbt!]
    \centering

    \begin{subfigure}[t]{0.48\columnwidth}
        \centering
        \includegraphics[width=\linewidth]{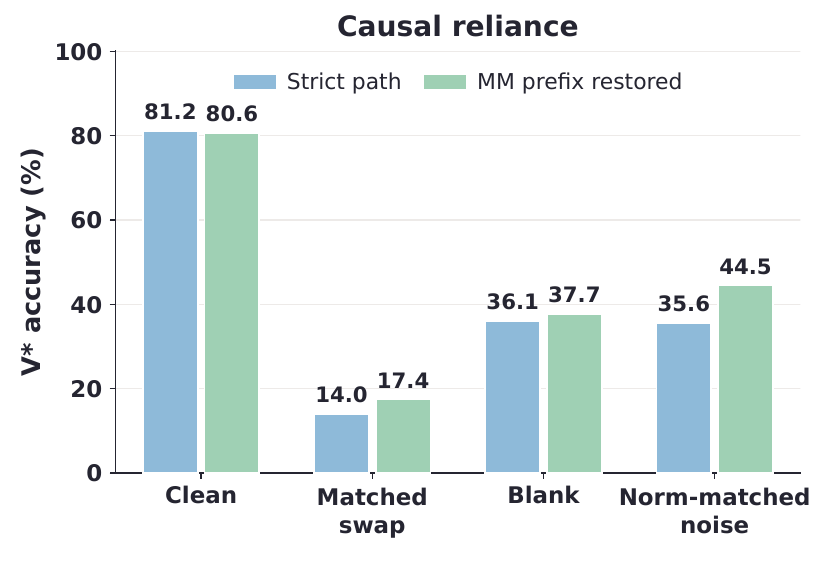}
        \caption{}
        \label{fig:causal_reliance}
    \end{subfigure}
    \hfill
    \begin{subfigure}[t]{0.48\columnwidth}
        \centering
        \includegraphics[width=\linewidth]{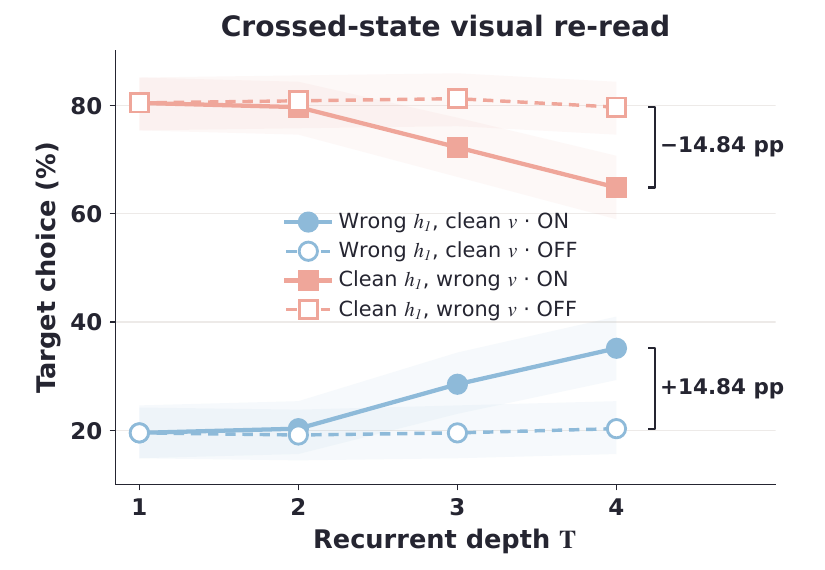}
        \caption{}
        \label{fig:crossed_reread}
    \end{subfigure}

    \caption{
Causal tests of recurrent computation.
(a) Corrupting $h_T$ degrades prediction under the strict path, while restoring the multimodal prefix partially compensates for the corrupted state.
(b) Clean $v$ corrects an incorrect $h_1$, while incorrect $v$ degrades a clean $h_1$ when question-to-visual KV connections are enabled (ON); blocking these connections (OFF) largely removes these effects.
}
    \label{fig:causal_recurrence}
\end{figure}

We first verify the claim enforced by the architecture: prediction under the strict interface should depend on recurrent content. Figure~\ref{fig:causal_reliance} shows that accuracy falls from $81.2\%$ to $14.0\%$ under matched recurrent-state replacement and to $36.1\%$ and $35.6\%$ under blank and norm-matched noise interventions. This sensitivity verifies path necessity, but it does not by itself establish visually grounded recurrent computation. Restoring the original multimodal prefix raises matched-swap and noise accuracy to $17.4\%$ and $44.5\%$. An alternative image-conditioned route can therefore compensate for corrupted recurrent content, directly illustrating why an informative latent state can become behaviorally nonessential when another visual path remains available.

We then test whether recurrence actually uses the persistent visual evidence. We construct crossed states that pair the question state from one image with visual evidence from its matched alternative. If $v$ causally guides the recurrent update, clean evidence should pull an incorrect $h_1$ toward the correct prediction, while incorrect evidence should pull a correct $h_1$ away from it. Figure~\ref{fig:crossed_reread} shows this bidirectional pattern when question-to-visual KV connections remain available, whereas blocking those connections nearly removes both effects, yielding difference-in-differences effects of $+14.8$ and $-14.8$ points. Thus, the recurrent trajectory is not merely a repeated transformation of $h_1$. It is causally revised by the visual evidence that remains available at each step.

\subsection{Image-Conditioned Content of the Recurrent State}
\label{sec:question_controlled}

\begin{wrapfigure}{r}{0.50\columnwidth}
    \centering
    \vspace{-1.5em}
    \includegraphics[width=0.5\columnwidth]{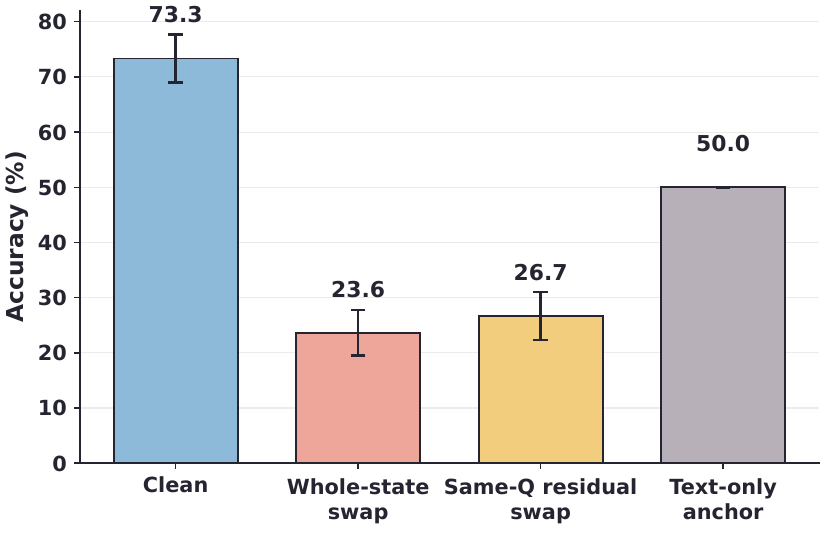}
\caption{
Same-question MMVP intervention with a fixed question and text-only anchor.
}
    \label{fig:question_controlled_swap}
    \vspace{-1.5em}
\end{wrapfigure}

A further alternative explanation remains. Replacing $h_T$ across examples changes both image-conditioned content and the question representation, so the degradation could reflect question disruption rather than visual content. We therefore use MMVP pairs in which two images $I$ and $I'$ share the same question $q$. For a target $(I,q)$, we replace $h_T(I,q)$ with $h_T(I',q)$ at decoding time. Because the question and token alignment remain fixed, this intervention isolates changes in image-conditioned content.

As shown in Figure~\ref{fig:question_controlled_swap}, clean accuracy is $73.3\%$, while whole-state and same-question replacement reduce it to $23.6\%$ and $26.7\%$. Similar sensitivity after fixing the question rules out question disruption as the primary explanation. The text-only anchor reaches $50.0\%$, above the $26.7\%$ obtained with content from the competing image, suggesting that incompatible image-conditioned content is more disruptive than removing such content alone. Together, these controls show that prediction follows the image-conditioned content carried by $h_T$ rather than merely requiring an intact hidden state.

\subsection{Step-Varying Visual Influence}
\label{sec:spatial_rereading}

\begin{figure}[t]
    \centering
    \includegraphics[width=\columnwidth]{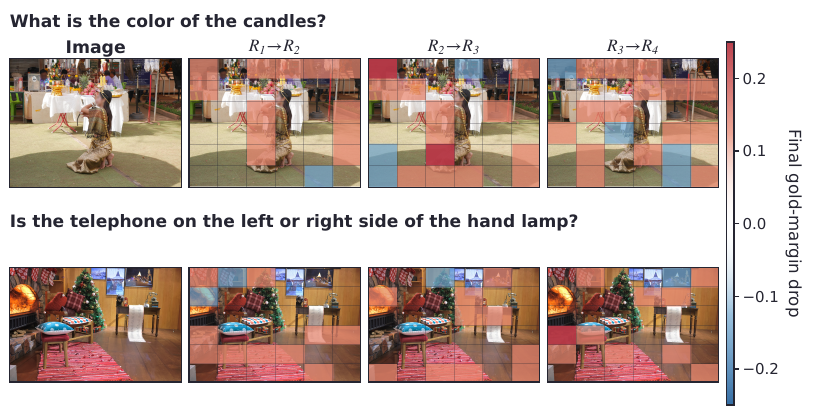}
    \caption{
Step-varying causal visual influence across recurrence.
Blocking local question-to-visual connections reveals image regions that affect the final gold-answer margin at each step.
}
    \label{fig:spatial_reread}
\end{figure}

The preceding interventions establish that recurrent updates use visual evidence, but not whether successive steps rely on the same evidence or shift which regions matter to prediction. We therefore block question-to-visual connections within each $6\times6$ image region and measure the resulting change in the final gold-answer margin. Figure~\ref{fig:spatial_reread} shows that causally influential regions shift across recurrent steps and, in the visualized examples, later transitions qualitatively emphasize objects implicated by the query. This variation also appears across the full $V^*$ set, where consecutive causal maps have mean cosine similarities of $0.37$ and $0.35$. Recurrence therefore does not treat $v$ as a static packet of visual information whose contribution remains fixed. Instead, as the question state evolves, different parts of the persistent evidence become causally relevant to the prediction. Full population statistics are reported in Appendix~\ref{app:spatial_statistics}. Taken together, the analyses support a coherent computational picture: $h_1$ carries pretrained visual competence into the required path, learned recurrence improves on that starting state, persistent visual evidence causally redirects the trajectory, and the visual evidence relevant to prediction changes across recurrent steps.
\section{Conclusion \& Limitations}

We address a central ambiguity in latent visual reasoning: a latent state can contain task-relevant visual information without being necessary for prediction. We introduce CVRR to align these properties by preserving pretrained visual competence while requiring image-conditioned prediction to pass through recurrent latent computation. CVRR starts from the native multimodal question state, refines it against persistent visual evidence, and removes alternative multimodal answer routes before decoding. Under this strict interface, CVRR retains strong visual reasoning performance, while causal analyses show that predictions follow the image-conditioned recurrent state and that persistent visual evidence revises its trajectory. These results support latent reasoning in which informativeness alone is insufficient and the proposed computation should also be a path that prediction actually uses. The visual-read boundary remains backbone-specific, our mechanistic analyses mainly use multiple-choice settings, and the current design relies on persistent rather than adaptively acquired visual evidence. Extending causal path necessity to more adaptive and open-ended multimodal reasoning remains an important direction.
\subsection*{Reproducibility statement}

To support reproducibility, we provide the implementation code and execution instructions for the key experiments in the supplementary material. These include the training and inference procedures for CVRR, recurrent reasoning settings, causal interventions, boundary localization, and the evaluation of the main baselines. Model and data configurations, training hyperparameters, inference settings, evaluation protocols, hardware, and other experimental details are documented in the appendix. We also distinguish the settings used for the main benchmarks from those used in the analysis experiments and describe the implementation conditions for the compared methods, including the bypass-free baselines. We verified the implementation and checked that the reported results are consistent with the corresponding experimental settings.
\subsection*{AI Use Statement}

Generative AI tools were used to assist with literature search and related-work discovery, as well as to improve wording and grammar in the manuscript. All AI-assisted suggestions, including references and literature-related information, were independently checked and verified by the authors. The research ideas, methodological design, experimental execution, result interpretation, and scientific claims were developed and verified by the authors. The authors take full responsibility for the content of the paper.

\subsubsection*{Acknowledgments}
We thank Jungwoo Park and Yein Park for their insightful feedback and discussions. This work was supported by the Ministry of Health \& Welfare, Republic of Korea (HR20C002103); the National Research Foundation of Korea (NRF) grants funded by the Korea government (MSIT) (RS-2023-00262002 and RS-2026-25510253); the ICT Creative Consilience Program through the Institute of Information \& Communications Technology Planning \& Evaluation (IITP) grant funded by the Korea government (MSIT) (IITP-2026-RS-2020-II201819); the Korea Health Technology R\&D Project through the Korea Health Industry Development Institute (KHIDI), funded by the Ministry of Health \& Welfare, Republic of Korea (RS-2025-25462758); and the ``Advanced GPU Utilization Support Program'' funded by the Government of the Republic of Korea (Ministry of Science and ICT).

\bibliography{iclr2027_conference}
\bibliographystyle{iclr2027_conference}

\clearpage
\appendix

\section{Related Works}
\label{app:related_work}

\subsection{Visual and Latent Visual Reasoning}
\label{app:related_work_LVR}

Recent visual reasoning methods improve inference by explicitly acquiring, revisiting, or manipulating visual evidence. VL-Rethinker encourages multimodal self-reflection through reinforcement learning~\citep{r1}, DeepEyes repeatedly searches for image regions relevant to the current reasoning step~\citep{r2}, and PixelReasoner incorporates pixel-space operations into the reasoning trajectory~\citep{r3}. In parallel, latent visual reasoning moves intermediate computation from explicit text into continuous hidden representations. LVR iteratively develops latent visual states~\citep{i6}, Monet uses continuous visual embeddings as intermediate reasoning representations~\citep{r4}, and SkiLa combines textual and latent computation~\citep{r5}. Laser explores richer latent representations~\citep{r6}, UniVLR integrates textual and visual evidence in a shared latent space~\citep{r7}, HyLaR develops a hybrid latent reasoning framework~\citep{r8}, and PEARL learns predictive latent representations from visual reasoning trajectories~\citep{r9}. Despite their architectural differences, these methods primarily study how latent states are constructed, updated, or coupled with visual evidence. Importantly, introducing a latent reasoning state does not by itself make that state necessary for prediction. When answer decoding retains access to the original image-conditioned hidden states, visual tokens, or multimodal prefix KV caches, image information can reach the answer through both the proposed latent state and the original multimodal pathway. The latter forms a bypass around the latent computation. Such a bypass need not harm predictive performance, but it makes strong benchmark accuracy insufficient to establish that the model actually reasons through the proposed latent state. CVRR instead focuses on controlling this computational interface while preserving the visual competence of the pretrained model.

\subsection{Causal Utilization in Multimodal and Latent Reasoning}

A related line of work distinguishes information encoded in an internal representation from information that actually influences model behavior. Causal mediation analyses show that changes in visual or intermediate representations need not propagate to the final prediction~\citep{i17}, while latent replacement and removal experiments find that predictions can remain stable even when intermediate states are substantially altered~\citep{r10,r12}. Mechanistic studies further separate the contribution of latent content from interface, boundary, or formatting effects~\citep{r11}. Related concerns arise in multimodal question answering, where strong predictions can partly reflect linguistic priors, answer-choice regularities, or other signals that reduce dependence on the intended visual evidence~\citep{r14,r15}. In latent visual reasoning, an analogous problem arises when the decoder can still attend to image-conditioned representations that were created before or alongside the latent trajectory. In this case, the latent state may remain highly informative or even improve prediction while being behaviorally nonessential because the original multimodal context provides a parallel route to the answer. Causal interventions are therefore useful for distinguishing latent informativeness from latent utilization, but post-hoc diagnosis alone does not guarantee that the intended path is used. CVRR turns this distinction into an architectural constraint by retaining the native image-conditioned initialization, allowing recurrent states to re-read persistent visual evidence, and removing alternative image-conditioned routes before decoding so that the final recurrent state becomes the required visual path to prediction.

\section{Implementation Details}
\label{app:implementation}

\subsection{CVRR Training and Recurrent Depth}

Unless otherwise specified, we use Qwen2.5-VL-7B-Instruct as the base multimodal backbone. We localize the causal boundary using training examples only, without access to any evaluation benchmark examples. This selects $\ell^*=20$, and we reuse layer $L_{21}$ as the shared recurrent transition with $T=4$ fixed as the default recurrent depth before downstream benchmark evaluation. The recurrent state is initialized from the native image-conditioned question representation $h_1$, while the visual state $v$ is retained as fixed evidence throughout recurrence. For text-only controls, we define $B$ by passing the same question without the image through the frozen lower backbone and the native recurrent layer $f_0$. This places $B$ at the same depth and token alignment as $h_1$ while removing image-conditioned content. Before decoding, visual rows and the original multimodal KV cache are removed. The answer decoder receives $h_T$ together with image-free text-only prefix caches, making $h_T$ the sole source of image-conditioned information. We cap the visual input at 8,192 visual tokens. The pretrained backbone remains frozen, and we adapt only the recurrent transition using LoRA with rank $32$, scaling factor $\alpha=16$, and dropout $0.01$. This results in $2{,}883{,}584$ trainable parameters out of approximately $8.295$B total parameters, or $0.0348\%$.

CVRR training minimizes answer-token cross entropy only. We use $\beta=0.50$ during both training and inference. We optimize with AdamW using a learning rate of $2\times10^{-5}$, a cosine learning-rate schedule, a warmup ratio of $0.05$, and gradient clipping at $1.0$. Models are trained for three epochs with a per-device batch size of $32$, gradient accumulation of $1$, bfloat16 precision, and random seed $0$ on four NVIDIA B200 180GB GPUs, giving an effective global batch size of $128$. We select the checkpoint with the lowest loss for evaluation. For recurrent-depth experiments, we independently train models for $T\in\{2,3,4,6,8\}$ under the same settings. The $T=1$ condition has no recurrent transition or trainable recurrent LoRA and directly decodes $h_1$ through the strict interface. All trainable component ablations in Table~\ref{tab:component_ablation} are independently retrained under the same recipe as Full CVRR, changing only the indicated component.

\subsection{No-Bypass Baselines}

For the no-bypass comparison, we use the six latent reasoning baselines for which the strict answer interface can be imposed while preserving the method's original latent pathway. For each compatible baseline, we retain its architecture, latent mechanism, and method-specific training objectives, including auxiliary losses when applicable. We standardize the Visual CoT data and common SFT settings, modifying only the answer interface to remove visual states and multimodal prefix KV caches before decoding. PEARL is excluded from this comparison because imposing the same strict decoder interface would require altering its predictive latent pathway rather than only removing an alternative multimodal answer route. Thus, the comparison tests whether existing latent mechanisms can retain visual competence when alternative image-conditioned routes are removed without redefining those mechanisms to fit CVRR's interface.

\subsection{Evaluation Protocol}

We use greedy decoding for CVRR and all baselines and otherwise retain each implementation's default generation configuration, including its maximum-new-token setting. The main benchmark results and preliminary latent-replacement experiments use generation-based evaluation. Unless otherwise specified, the causal and mechanistic analyses use restricted first-option-token readout. Each trained model is evaluated from a single training run with random seed $0$. Reported 95\% confidence intervals are computed by nonparametric bootstrap over evaluation examples and quantify evaluation-sample uncertainty rather than decoding or training-run variability. For MMVP, bootstrap resampling is clustered by image pair to preserve within-pair dependence.

\section{Algorithm}
\label{app:algorithm}

Algorithm~\ref{alg:cvrr} summarizes the complete training and inference procedure of CVRR. The native multimodal forward first initializes the image-conditioned question state and persistent visual evidence. Recurrence then updates only the question state while repeatedly re-reading the fixed visual evidence through a shared adapted layer. Before answer decoding, all visual states and multimodal caches are discarded, leaving the final recurrent state as the only image-conditioned input to the decoder.

\begin{algorithm}[hbt!]
\caption{Training and inference with CVRR}
\label{alg:cvrr}
\begin{algorithmic}[1]

\Require Frozen VLM $M=\{L_1,\ldots,L_N\}$ and causal boundary $\ell^\star$
\Require Image--question pair $(I,q)$, recurrent depth $T$, mixing coefficient $\beta$
\Require Shared LoRA parameters $\phi$ on recurrent layer
         $F=L_{\ell^\star+1}$
\Ensure Answer distribution $p_\phi(y\mid I,q)$

\Statex
\Statex \textit{\textbf{Native multimodal initialization}}

\State $H_{\mathrm{mm}}^\star
    \gets M_{\leq\ell^\star}^{\mathrm{mm}}(I,q)$

\State $(Q_{\mathrm{txt}}^\star,
        \mathcal K_{\mathrm{txt}}^{\leq\ell^\star})
    \gets
    M_{\leq\ell^\star}^{\mathrm{txt}}
    (q;\mathtt{cache}=\mathrm{on})$

\State Disable $\phi$ and denote the native recurrent layer by $F_0$

\State $H_1
    \gets
    F_0(H_{\mathrm{mm}}^\star;
        \mathtt{cache}=\mathrm{off})$

\State $(B,\mathcal K_{\mathrm{txt}}^{\ell^\star+1})
    \gets
    F_0(Q_{\mathrm{txt}}^\star;
        \mathtt{cache}=\mathrm{on})$
    \Comment{text-only anchor at the $h_1$ depth}

\State $\mathcal K_{\mathrm{txt}}
    \gets
    \bigl(
    \mathcal K_{\mathrm{txt}}^{\leq\ell^\star},
    \mathcal K_{\mathrm{txt}}^{\ell^\star+1}
    \bigr)$

\State $h_1 \gets P_Q H_1$
\State $v \gets P_V H_1$
    \Comment{persistent visual evidence}

\State Enable the shared recurrent adapter $\phi$

\Statex
\Statex \textit{\textbf{Persistent visual re-reading}}

\For{$t=2,\ldots,T$}

    \State $X_t
        \gets
        \operatorname{Replace}_Q(H_1,h_{t-1})$

    \Statex \hspace{\algorithmicindent}
        $\triangleright$ Keep $P_VX_t=v$ and preserve the native
        mask, token ordering, and positions

    \State $\widetilde h_t
        \gets
        P_Q F_\phi
        (X_t;\mathtt{cache}=\mathrm{off})$

    \State $h_t
        \gets
        (1-\beta)h_{t-1}
        +\beta\widetilde h_t$

\EndFor

\Statex
\Statex \textit{\textbf{Strict answer decoding}}

\State Discard $v$, $H_{\mathrm{mm}}^\star$, and all multimodal/recurrent caches

\State $p_\phi(y\mid I,q)
    \gets
    \operatorname{AnswerDecode}
    (h_T,\mathcal K_{\mathrm{txt}})$

\Statex \hspace{\algorithmicindent}
    $\triangleright$ $h_T$ is the only image-conditioned state
    available to the answer decoder

\If{training with gold answer $y$}

    \State $\displaystyle
    \ell_i
    \gets
    -\frac{1}{|\mathcal A_i|}
    \sum_{j\in\mathcal A_i}
    \log
    p_\phi
    (y_{i,j}\mid
     y_{i,<j},
     h_{T,i},
     \mathcal K_{\mathrm{txt},i})$

    \State $\displaystyle
    \mathcal L_{\mathrm{ans}}
    \gets
    \frac{1}{|\mathcal B|}
    \sum_{i\in\mathcal B}\ell_i$

    \State Update only $\phi$ using
    $\nabla_\phi\mathcal L_{\mathrm{ans}}$

\EndIf

\State \Return $p_\phi(y\mid I,q)$

\end{algorithmic}
\end{algorithm}

\section{Efficiency and Computational Cost}
\label{app:efficiency}

\begin{table}[hbt!]
\centering
\caption{
Accuracy and inference cost across recurrent depths.
Latency is measured on a single NVIDIA B200 over 32 examples with three repeated runs, excluding image preprocessing.
Relative latency is normalized to $T=1$.
FLOPs report decoder-side computation, and peak $\Delta$ memory reports incremental GPU memory during inference.
}
\label{tab:efficiency_depth}
\small
\setlength{\tabcolsep}{4.5pt}
\renewcommand{\arraystretch}{1.08}
\begin{tabular}{cccccc}
\toprule
$T$
& $V^*$ Acc. (\%)
& Latency (s)
& Rel. Latency
& TFLOPs
& Peak $\Delta$ Mem. (GiB) \\
\midrule
1 & 75.4 & 1.044 & 1.000 & 47.58 & 1.04 \\
2 & 78.5 & 1.099 & 1.053 & 49.70 & 1.12 \\
3 & 79.6 & 1.163 & 1.114 & 51.83 & 1.16 \\
\textbf{4} & \textbf{81.2} & 1.146 & 1.098 & 53.96 & 1.16 \\
6 & 78.5 & 1.178 & 1.129 & 58.22 & 1.16 \\
8 & 80.1 & 1.154 & 1.106 & 62.48 & 1.16 \\
\bottomrule
\end{tabular}
\end{table}

Table~\ref{tab:efficiency_depth} reports how inference cost and accuracy change with recurrent depth. The $T=1$ diagnostic reaches $75.4\%$ on $V^*$, while independently trained recurrent models reach $78.5\%$ at $T=2$ and peak at $81.2\%$ at the default $T=4$. The trend is not monotonic: accuracy decreases to $78.5\%$ at $T=6$ and recovers to $80.1\%$ at $T=8$. Because models with $T\geq2$ are independently trained at their respective depths, this sweep should not be interpreted as a monotonic test-time scaling law. Instead, it shows that additional recurrent depth is useful only up to a task- and training-dependent regime, with $T=4$ providing the best accuracy among the tested settings. From $T=1$ to $T=4$, measured latency increases by $9.9\%$ and decoder-side computation by $13.4\%$, while peak incremental memory increases by only $0.12$ GiB.

\begin{table}[hbt!]
\centering
\caption{
Inference cost compared with six latent visual reasoning baselines under a matched four-step budget.
Latency is measured on a single NVIDIA B200 over 32 examples with three repeated runs, excluding image preprocessing.
Peak and $\Delta$ report total and incremental GPU memory during inference.
}
\label{tab:efficiency_baselines}
\small
\setlength{\tabcolsep}{6.0pt}
\renewcommand{\arraystretch}{1.08}
\begin{tabular}{lccc}
\toprule
Model
& Latency (s)
& Decoder TFLOPs
& Peak / $\Delta$ (GiB) \\
\midrule
\textbf{CVRR}
& \textbf{$1.146 \pm 0.114$}
& \textbf{53.96}
& \textbf{16.62 / 1.16} \\
LVR-7B
& $1.397 \pm 0.091$
& 59.72
& 17.38 / 1.92 \\
Monet-7B
& $1.396 \pm 0.092$
& 59.85
& 17.30 / 1.85 \\
SkiLa-7B
& $1.377 \pm 0.090$
& 59.70
& 17.38 / 1.92 \\
Laser-7B
& $1.329 \pm 0.085$
& 59.70
& 17.31 / 1.86 \\
UniVLR-7B
& $1.516 \pm 0.090$
& 59.70
& 18.38 / 2.88 \\
HyLaR-7B
& $1.378 \pm 0.090$
& 59.70
& 17.38 / 1.92 \\
\bottomrule
\end{tabular}
\end{table}

Table~\ref{tab:efficiency_baselines} compares CVRR with six latent visual reasoning baselines under the same four-step reasoning budget. CVRR has the lowest measured latency, decoder-side computation, and incremental memory among these models. Its latency is $1.146$ seconds compared with $1.329$--$1.516$ seconds for the baselines, while decoder computation is $53.96$ TFLOPs compared with approximately $59.7$--$59.9$ TFLOPs. CVRR also uses $1.16$ GiB of incremental memory compared with $1.85$--$2.88$ GiB for the baselines. This efficiency follows from reusing a single native decoder layer across recurrent steps rather than repeatedly invoking a larger portion of the multimodal backbone. We use $T=4$ throughout the main experiments unless otherwise specified.

\section{Query-Channel Corrections Across Recurrence}
\label{app:query_corrections}

\begin{figure}[hbt!]
    \centering
    \includegraphics[width=\columnwidth]{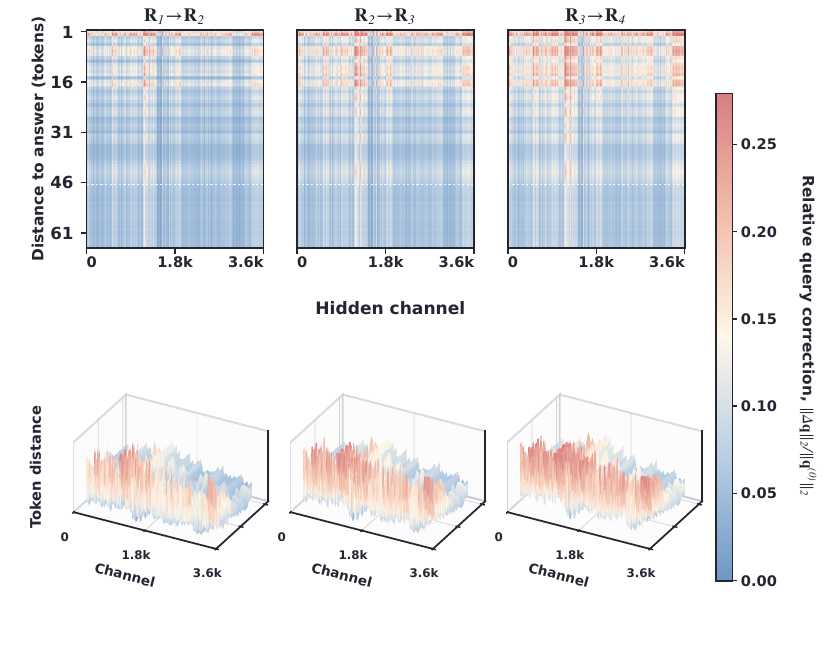}
    \caption{
    Query-channel corrections across recurrent transitions.
    Top panels show relative correction magnitude across question-token positions and hidden channels for $R_1\!\rightarrow\!R_2$, $R_2\!\rightarrow\!R_3$, and $R_3\!\rightarrow\!R_4$.
    Bottom panels show the corresponding three-dimensional surfaces.
    }
    \label{fig:query_channel_correction}
\end{figure}

We further examine how recurrent computation modifies the question representation across successive visual re-reads. We report the relative query correction $\|\Delta q\|_2/\|q^{(0)}\|_2$, where $q^{(0)}$ denotes the corresponding reference query representation and $\Delta q$ denotes the recurrent correction. Figure~\ref{fig:query_channel_correction} visualizes this quantity across question-token positions and hidden channels for each recurrent transition. The aggregate correction increases from $8.9\%$ at $R_1\!\rightarrow\!R_2$ to $10.6\%$ at $R_2\!\rightarrow\!R_3$ and $12.0\%$ at $R_3\!\rightarrow\!R_4$. Later recurrent steps therefore continue to modify the question state rather than simply reproducing the initial multimodal representation.

The correction is non-uniform across both token positions and hidden channels. The two-dimensional maps show this structured variation directly, while the three-dimensional surfaces provide a complementary view of its magnitude and concentration. These quantities characterize functional changes induced by recurrent updates and should not be interpreted as attention weights or semantic feature importance. We avoid strong conclusions from the farthest token-distance regions because the number of available examples decreases substantially in the tail.

\section{Projection-Wise Functional Corrections}
\label{app:projection_correction}

\begin{figure}[hbt!]
    \centering
    \includegraphics[width=\columnwidth]{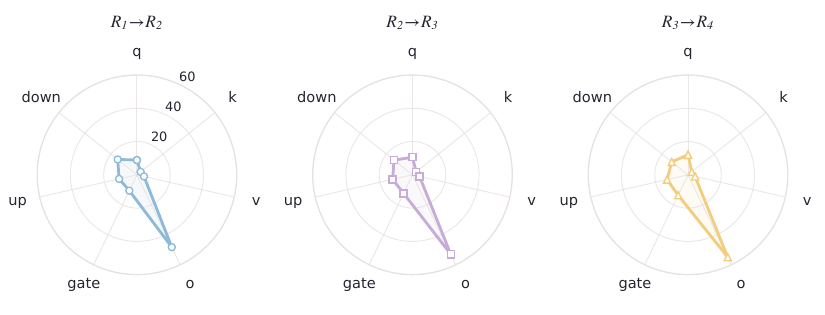}
    \caption{
    Projection-wise functional corrections across recurrence.
    Polar plots show relative correction magnitude for transformer projections at $R_1\!\rightarrow\!R_2$, $R_2\!\rightarrow\!R_3$, and $R_3\!\rightarrow\!R_4$.
    All panels use the same projection ordering and radial scale.
    }
    \label{fig:projection_correction}
\end{figure}

\begin{table}[hbt!]
\centering
\caption{
Projection-wise functional correction magnitudes (\%).
}
\label{tab:projection_correction}
\small
\setlength{\tabcolsep}{4.5pt}
\begin{tabular}{lccc}
\toprule
Projection
& $R_1\!\rightarrow\!R_2$
& $R_2\!\rightarrow\!R_3$
& $R_3\!\rightarrow\!R_4$ \\
\midrule
$q_{\mathrm{proj}}$    & 8.9  & 10.6 & 12.0 \\
$k_{\mathrm{proj}}$    & 2.9  & 2.9  & 2.9  \\
$v_{\mathrm{proj}}$    & 4.3  & 4.3  & 4.3  \\
$o_{\mathrm{proj}}$    & \textbf{48.3} & \textbf{53.0} & \textbf{54.9} \\
$\mathrm{gate}_{\mathrm{proj}}$ & 10.6 & 12.4 & 13.6 \\
$\mathrm{up}_{\mathrm{proj}}$   & 11.0 & 12.3 & 12.9 \\
$\mathrm{down}_{\mathrm{proj}}$ & 14.8 & 14.2 & 12.5 \\
\bottomrule
\end{tabular}
\end{table}

We further decompose the recurrent update by transformer projection to examine where functional corrections are expressed within the shared recurrent layer. Figure~\ref{fig:projection_correction} and Table~\ref{tab:projection_correction} show that the output projection receives the largest relative correction throughout recurrence, increasing from $48.3\%$ to $53.0\%$ and $54.9\%$ across the three transitions. Corrections in the query, gate, and up projections also increase across recurrent steps, while the key and value projections remain comparatively stable.

This pattern is consistent with the CVRR design, where visual evidence remains fixed while the question representation evolves across recurrent steps. The query- and output-side corrections show that substantial modification continues to be expressed in the evolving recurrent state, while key and value corrections vary little across transitions. These quantities characterize where functional corrections are expressed within the shared layer rather than the semantic importance of individual projections, and should not be interpreted as showing that any single projection uniquely performs visual reasoning.

\section{Population Statistics for Spatial Visual Re-reading}
\label{app:spatial_statistics}

Table~\ref{tab:spatial_reread} reports population-level statistics over the full $V^*$ evaluation set to complement the qualitative examples in Figure~\ref{fig:spatial_reread}. Causal influence is spatially concentrated across all recurrent transitions, while consecutive region maps show only moderate similarity. These results support that the step-varying visual re-reading observed in the examples is not driven by a small set of selected cases.

\begin{table}[hbt!]
\centering
\caption{
Population statistics for spatial causal re-reading on $V^*$.
Intervals denote 95\% confidence intervals.
}
\label{tab:spatial_reread}
\small
\setlength{\tabcolsep}{3.5pt}
\resizebox{\columnwidth}{!}{%
\begin{tabular}{lccc}
\toprule
Statistic
& $R_1\!\rightarrow\!R_2$
& $R_2\!\rightarrow\!R_3$
& $R_3\!\rightarrow\!R_4$ \\
\midrule
Mean signed effect
& $0.008\;[-0.004,\,0.020]$
& $0.006\;[-0.006,\,0.018]$
& $0.005\;[-0.007,\,0.017]$ \\

Max positive effect
& $0.113\;[0.099,\,0.127]$
& $0.120\;[0.107,\,0.133]$
& $0.111\;[0.098,\,0.125]$ \\

Positive regions (\%)
& $24.6\;[20.8,\,28.7]$
& $24.3\;[20.4,\,28.2]$
& $24.0\;[20.3,\,27.9]$ \\

Top-quartile positive mass (\%)
& $48.4\;[42.6,\,54.2]$
& $52.9\;[47.1,\,58.8]$
& $46.6\;[40.9,\,52.5]$ \\

Next-map cosine similarity
& $0.367\;[0.327,\,0.407]$
& $0.350\;[0.311,\,0.390]$
& -- \\
\bottomrule
\end{tabular}%
}
\end{table}

\section{Localizing the Causal Visual-Read Interface}
\label{app:boundary_localization}

\begin{figure}[hbt!]
    \centering

    \begin{subfigure}[t]{0.5\columnwidth}
        \centering
        \includegraphics[width=\linewidth]{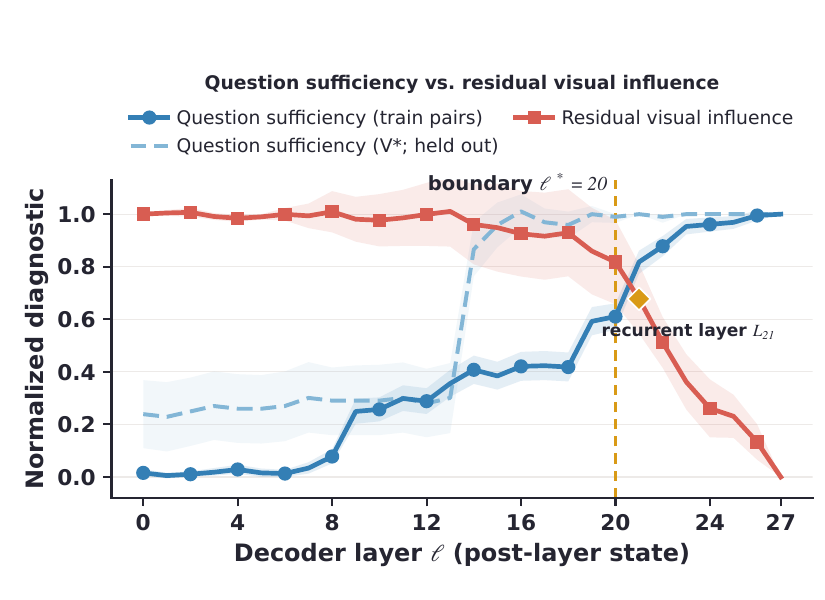}
        \caption{}
        \label{fig:boundary_tradeoff}
    \end{subfigure}
    \hfill
    \begin{subfigure}[t]{0.49\columnwidth}
        \centering
        \includegraphics[width=\linewidth]{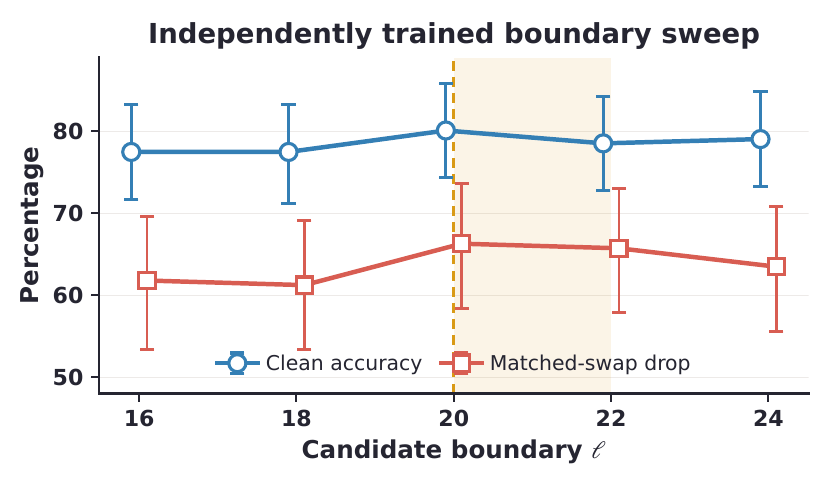}
        \caption{}
        \label{fig:boundary_sweep}
    \end{subfigure}

    \caption{
    Localizing the causal visual-read interface.
    (a) Question-state sufficiency and residual visual influence characterize the selected transition near $\ell^*=20$; the dashed $V^*$ curve is held out from boundary selection.
    (b) Independent boundary training supports the resulting $20$--$22$ region.
    }
    \label{fig:boundary_localization}
\end{figure}

Selecting a recurrent boundary requires more than locating where visual mediation begins to decline. The boundary is selected from training-pair activation-patching profiles using the first sustained decline criterion described in the main text. To interpret and validate this choice, we additionally examine whether the selected region already provides a question state with substantial pretrained visual competence. Question-state sufficiency evaluates strict decoding from the native post-layer question state after removing visual rows and multimodal caches, while residual visual influence measures the effect of visual-row intervention on the downstream answer.

Figure~\ref{fig:boundary_tradeoff} shows why the selected transition is suitable for recurrence. Early layers retain strong residual visual influence but provide insufficient question states for strict decoding. With increasing depth, question-state sufficiency rises while residual visual influence eventually declines, creating a region where a competent question state coexists with visual signal that remains available for further refinement. The activation-patching criterion selects $\ell^*=20$, immediately before recurrent layer $L_{21}$. The training-pair sufficiency curve supports this choice, while the dashed $V^*$ curve is held out from selection and independently confirms that question-state sufficiency is already high in this region.

We further validate the selected region by training CVRR at nearby candidate boundaries under the same training budget. Figure~\ref{fig:boundary_sweep} shows that layers 20 and 22 achieve $80.10\%$ and $78.53\%$ clean accuracy with matched-swap drops of $66.29$ and $65.73$ points. Together, the sufficiency--influence analysis and boundary sweep support the $20$--$22$ region as a functional visual-read interface rather than an insertion point chosen from downstream benchmark accuracy.

\section{Generalization Across Multimodal Backbones}
\label{app:backbone_generalization}

\begin{table}[hbt!]
\centering
\caption{
Generalization across multimodal backbones.
Strict--Full reports the paired accuracy difference, and intervention columns report drops under recurrent-content corruption.
}
\label{tab:backbone_generalization}
\scriptsize
\setlength{\tabcolsep}{2.3pt}
\resizebox{\columnwidth}{!}{%
\begin{tabular}{lccccccc}
\toprule
Backbone
& $\ell^*$
& Full MM
& Strict
& $\text{Strict}-\text{Full}$ (95\% CI)
& Swap drop
& Blank drop
& Noise drop \\
\midrule
Gemma-3-4B
& 21
& 36.1
& 40.3
& +4.2 [-3.1, +11.5]
& 7.3
& 9.4
& 9.4 \\

Gemma-3-12B
& 25
& 43.5
& 44.0
& +0.5 [-5.2, +6.3]
& 12.4
& 11.5
& 10.5 \\

Gemma-4-12B
& 32
& 52.9
& 48.2
& -4.7 [-10.5, +0.5]
& 20.2
& 14.7
& 26.7 \\

InternVL-3-9B
& 35
& 75.4
& 75.9
& +0.5 [-1.1, +2.6]
& 57.1
& 42.9
& 42.4 \\
\bottomrule
\end{tabular}%
}
\end{table}

Finally, we test whether the competence--reliance pattern observed with Qwen2.5-VL transfers to other multimodal backbones. For each additional model, we independently localize its visual-read boundary and train the same strict CVRR interface. As shown in Table~\ref{tab:backbone_generalization}, InternVL3-9B~\citep{a3} reaches $75.9\%$ under the strict path compared with $75.4\%$ for the full multimodal continuation, while matched recurrent-content replacement causes a $57.1$ percentage-point drop. Gemma-3-12B~\citep{a1} and Gemma-4-12B~\citep{a2} also retain performance close to their full multimodal continuations while showing sensitivity to recurrent-content corruption. The effect is weaker for Gemma-3-4B, indicating that the strength of recurrent-state reliance varies across backbones. Overall, these results suggest that the combination of competence preservation and recurrent reliance transfers beyond Qwen2.5-VL, while its magnitude remains architecture-dependent.

\end{document}